\pdfoutput=1

\documentclass[11pt]{article}

\newcommand{\ie}{\textit{i}.\textit{e}.}
\newcommand{\eg}{\textit{e}.\textit{g}.}

\usepackage{naacl2021}
\usepackage{times}
\usepackage{latexsym}
\usepackage{multirow}
\usepackage{graphicx}
\usepackage{pifont}
\usepackage{xcolor,pifont}
\usepackage{tablefootnote}
\usepackage[export]{adjustbox}
\usepackage[T1]{fontenc}

\usepackage[utf8]{inputenc}

\usepackage{microtype}

\title{Mobile App Tasks with Iterative Feedback (MoTIF): \\Addressing Task Feasibility in Interactive Visual Environments 
}

\author{Andrea Burns\textsuperscript{\normalfont 1} \and Deniz Arsan\textsuperscript{\normalfont 2} \and Sanjna Agrawal\textsuperscript{\normalfont 1} \and Ranjitha Kumar\textsuperscript{\normalfont 2} \\
{\bf Kate Saenko\textsuperscript{\normalfont 1,3}} \and {\bf Bryan A. Plummer\textsuperscript{\normalfont 1}}\\
\textsuperscript{\normalfont 1}Boston University, MA \\
\texttt{\{aburns4,sanjna,saenko,bplum\}@bu.edu} \\
\textsuperscript{\normalfont 2}University of Illinois at Urbana-Champaign, IL \\
\texttt{\{darsan2,ranjitha\}@illinois.edu} \\
\textsuperscript{\normalfont 3}MIT-IBM Watson AI Lab, MA \\}

\begin{document}
\maketitle
\begin{abstract}
In recent years, vision-language research has shifted to study tasks which require more complex reasoning, such as interactive question answering, visual common sense reasoning, and question-answer plausibility prediction. However, the datasets used for these problems fail to capture the complexity of real inputs and multimodal environments, such as ambiguous natural language requests and diverse digital domains. 
We introduce Mobile app Tasks with Iterative Feedback (MoTIF), a dataset with natural language commands for the greatest number of interactive environments to date.\footnote{MoTIF's collection is ongoing and its current version can be found at \url{https://github.com/aburns4/MoTIF}.} 
MoTIF is the first to contain natural language requests for interactive environments that are not satisfiable, and we obtain follow-up questions on this subset to enable research on task uncertainty resolution. We perform initial feasibility classification experiments and only reach an F1 score of 37.3, verifying the need for richer vision-language representations and improved architectures to reason about task feasibility.
\end{abstract}
\begin{table*}[h]
\setlength{\tabcolsep}{2.5pt}
\renewcommand{\arraystretch}{0.9}
    \centering
    \begin{tabular}{l|c|c|c|c|c|c|c}
    \hline
        Dataset & Domain & \# Envs & \# NL Tasks & \# Views & Interactive & Real & Feasibility\\
        \hline
         MiniWoB~\cite{wob} & \multirow{2}{*}{Webpage} & 100 & 0 & 1 & \color{green}{\ding{51}} & \color{red}{\ding{55}} & \color{red}{\ding{55}}\\
         \citet{langtoelem} & & 1,800 & 50,000 & 1 & \color{red}{\ding{55}} & \color{green}{\ding{51}}& \color{red}{\ding{55}}\\
         \hline
         R2R~\cite{Anderson_2018_CVPR_NAV} & \multirow{4}{*}{House}& 90 & 21,567 & -- &  \color{green}{\ding{51}} &  \color{green}{\ding{51}} & \color{red}{\ding{55}}\\
         EQA~\cite{embodiedqa} & & 45,000 & 0 & -- &  \color{green}{\ding{51}} & \color{red}{\ding{55}}& \color{red}{\ding{55}}\\
         IQA~\cite{IQA} & & 30 & 0 & -- & \color{green}{\ding{51}} & \color{red}{\ding{55}} & \color{red}{\ding{55}}\\
         ALFRED~\cite{ALFRED20} & & 120 & 25,743 & -- &  \color{green}{\ding{51}} &\color{red}{\ding{55}}& \color{red}{\ding{55}}\\
         \hline
         Rico~\cite{rico} & \multirow{3}{*}{App} & 9,700 & 0 & 6.7 &  \color{red}{\ding{55}} &  \color{green}{\ding{51}}& \color{red}{\ding{55}}\\
         PIXELHELP~\cite{li-etal-2020-mapping} & & 4 & 187 & 4 &  \color{green}{\ding{51}} &  \color{green}{\ding{51}} & \color{red}{\ding{55}} \\
         \textbf{MoTIF} &  & 125+ & 6,100+ & 14 & \color{green}{\ding{51}} & \color{green}{\ding{51}}&  \color{green}{\ding{51}}\\
         \hline
    \end{tabular}
    \vspace{-2mm}
    \caption{Comparison of MoTIF to existing datasets. We consider the number of environments, natural language commands, and views, in addition to whether the environment is interactive, real (not simulated), and captures task feasibility. We provide the average number of views for Rico and MoTIF; PIXELHELP reports the median.}
    \label{tab:compare}
    \vspace{-2mm}
\end{table*}

\section{Introduction}

Vision-language tasks often require high level reasoning skills like counting, comparison, and common sense to relate visual and language data~\citep{IQA,zhang2019raven,Gardner2020DeterminingQP}. Prior works' abilities to learn and employ this form of reasoning has been shown to be neither reliable nor robust when used in realistic settings where there is task uncertainty or environment variation. Task infeasibility (when a task may not be possible) can cause vision-language models to generate visually unrelated, yet plausible answers~\citep{massiceti}.
This is dangerous for users that are limited in their ability to determine if an answer is trustworthy, either physically or situationally, \eg, users that are low-vision or driving. Vision-language models also often experience large performance drops in new environments due to domain shift, reducing the impact of prior work in application~\cite{Yu_2020_CVPR_Workshops}.
These are fundamental machine learning problems, and they begin with the data used to train and evaluate learned models.
\begin{figure}[t]
    \includegraphics[scale=0.0615,left]{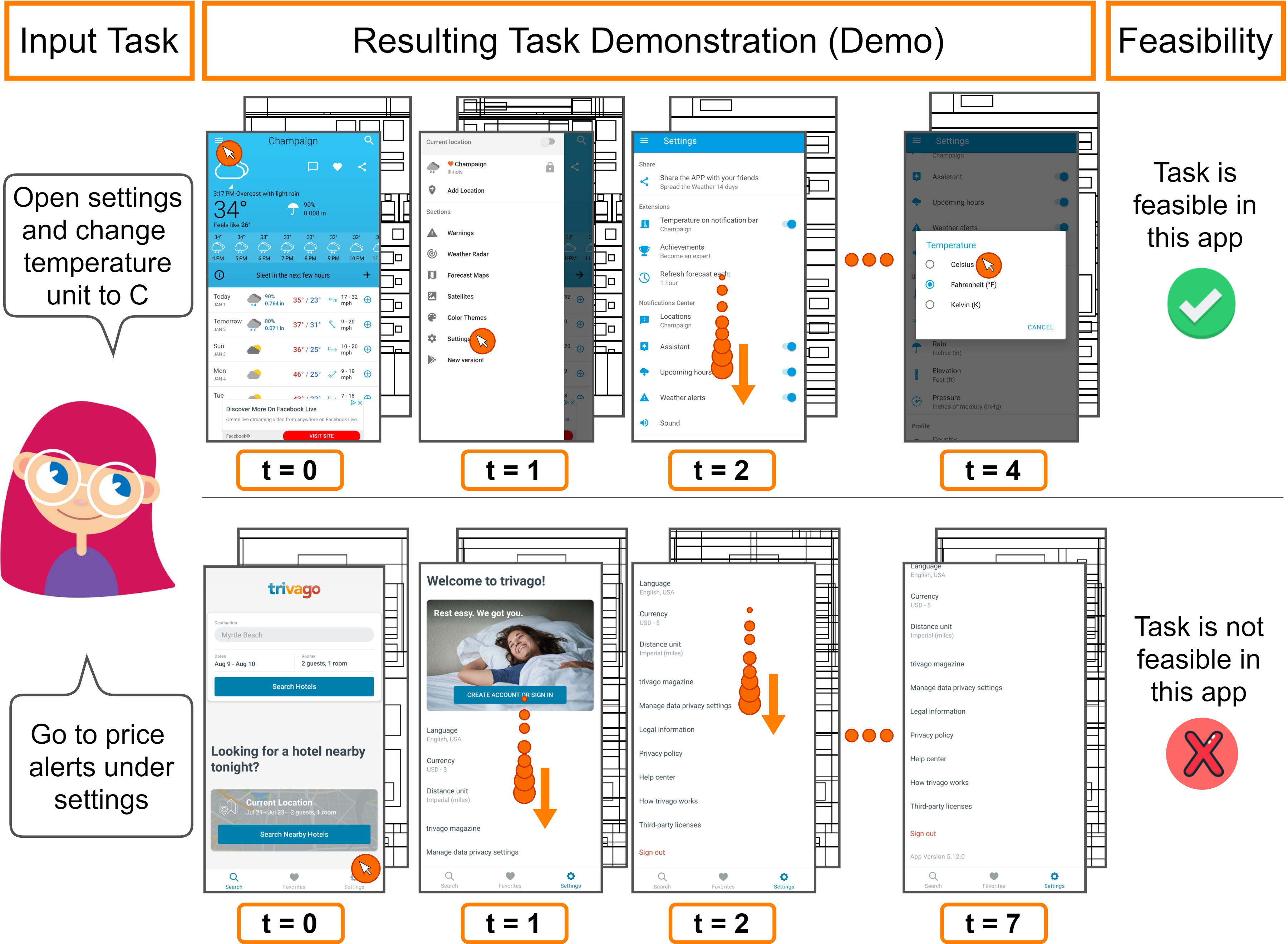}
    \caption{Example MoTIF tasks and their demos. Annotators attempt natural language tasks in apps. We obtain a demo of the attempt and find out if it was possible. For each time step, we capture action coordinates (\ie, where clicking, typing, or scrolling occurs) and the app screen and view hierarchy (illustrated behind it).}
    \label{fig:motivation}
    \vspace{-5mm}
\end{figure}


We propose Mobile app Tasks with Iterative Feedback (MoTIF), the first large scale dataset for interactive natural language app tasks. 
Mobile apps have a rich variety of environments with challenging decision landscapes, unlike current vision-language tasks which use well constrained images or simulated environments. Moreover, MoTIF focuses on goal-oriented tasks within apps, while current phone assistants and prior work are limited to voice commands for information retrieval or simple device-related commands~\cite{li-etal-2020-mapping}. 
MoTIF provides greater linguistic complexity for interactive tasks with over 6.1k free form natural language commands for tasks in 125 Android apps. 
Its task demos include the app view hierarchy, screen, and action coordinates for each time step, as shown in Figure~\ref{fig:motivation}. MoTIF uniquely includes binary feasibility annotations for each task, subclass annotations for why tasks are infeasible, and follow up questions. Data collection is ongoing; we have collected task demos for five tasks per app thus far.\footnote{We have collected demos for nearly 100 apps and decided to not collect demos for dating apps due to privacy reasons. We are resolving technical issues with the few remaining apps.}

We provide initial results for the simplified task of predicting a task command to be feasible or not. We leave multiclass classification of why a task is not possible and task automation to future work. We hope automating mobile app tasks and capturing realistic task infeasibility will enable users of all ability levels to engage with mobile apps with ease.
We also collect demos of the same task across multiple apps to encourage research in task generalization, so that resulting tools are robust to domain shift and ultimately higher impact in application.
\section{Related Work}

MoTIF subsumes several datasets and research topics: web task automation, vision-language navigation (VLN), task feasibility prediction, and app design; we provide a comparison in Table~\ref{tab:compare}.
Prior work in automating web tasks~\cite{wob,langtoelem} limit user interaction to a single screen, unlike MoTIF which contains task demonstrations with an average of 14 visited screens. Recently, PIXELHELP~\cite{li-etal-2020-mapping} was proposed as a small evaluation-only dataset for 187 natural language tasks in Pixel phones, but the majority are device specific (\ie, not in-app commands). As for VLN datasets, they tend to either have many natural language commands and few environments, or vice versa, and most use simulated environments.

Importantly, none of these prior works capture task infeasibility.
Vision-language research has recently begun to explore this topic: VizWiz~\cite{vizwiz} introduced a visual question answering dataset for images taken by people that are blind, resulting in questions which may not be answerable. To the best of our knowledge, VizWiz is the only vision-language dataset with task infeasibility, but it concerns static images. Additionally, images that cannot be used to answer visual questions are easily classified as such, as they often are blurred or contain random scenes (\eg, the floor).~\citet{Gardner2020DeterminingQP} explored question-answer plausibility prediction, but the questions used were generated from a bot, which could result in extraneous questions that are easy to classify as implausible. Both are significantly different from the nuanced tasks of MoTIF, for which exploration is necessary to determine task feasibility. Its infeasible tasks are always within the same Android app category, having an inherent relevance to the visual environment.

\section{Data Collection}
Apps were chosen over fifteen Google Play Store categories
ensuring each had at least 50k downloads and a rating of 4/5. We use UpWork to crowd source MoTIF and now detail how we collect task commands, demos, and feasibility annotations:

\smallskip
\noindent\textbf{Natural Language Commands}
We instruct workers to write tasks as if they are asking the app to perform the task for them. The annotators are free to explore the app before submitting their tasks. 
We neither structure the tasks nor prescribe a number of tasks to be written; 
this creates natural language tasks that mimic real users, unlike automatically generated tasks from prior work~\cite{wob}. 

\smallskip
\noindent\textbf{Task-Application Pairing}
We select an initial subset of tasks to collect demos for by clustering tasks within an Android app category. This captures realistic task infeasibility and we plan to extend MoTIF to all (task, app) combinations within each app category. We apply K-Means~\cite{kmeans} over the natural language tasks using the average FastText embedding~\cite{joulin2016bag}. 
For task clusters with reasonable app variance, we assign one task near each cluster's centroid to all apps within that category. Clustering is performed using $K=5$, as we collect demos for five tasks per app for now.

If an app's tasks are not distributed across clusters, we leave the (task, app) pairs \textit{app-specific}, or pair tasks with one to two other apps. App-specific refers to annotators having explored this app before submitting tasks for it during our task collection stage (as opposed to our clustered pairing). This resulted in 41 apps with category-clustered commands.
When analyzing feasibility annotations, we find that both app-specific and category-clustered (task, app) pairs contain infeasible tasks. 

\smallskip
\noindent\textbf{Task Demos \& Feasibility Annotations} Next, we provide annotators with instructions to complete the task in the provided app. 
Workers interact with Android devices remotely through a website that is reachable on any web browser and are provided anonymized information if needed for logging in. 
After attempting the task, they are brought to a post-survey to answer if they successfully completed the task, and if not, why. 
The survey contains multiple choice questions and fill-in the blank options regarding task feasibility detailed in Section~\ref{sec:analysis}.
\section{Data Analysis}
\label{sec:analysis}
We now analyze the collected natural language tasks, feasibility annotations, and task demos.

\smallskip
\noindent\textbf{Natural Language Commands}
We collected 6.1k natural language tasks over 125 Android apps.
After removing non-alphanumeric characters and stop words, the vocabulary size was 3,658 words
, with the average task length being 5.6 words. The minimum task length is one, consisting of single action commands like ``refresh'' or ``login,'' with the longest consisting of 44 words. Average task length has a range of 1.5 words over all categories.

\smallskip
\noindent\textbf{Feasibility Annotations}
Thus far, we collected up to ten demos for 480 (task, app) pairs, creating nearly 4.7k demos. Of the (task, app) pairs, 143 are deemed infeasible by at least five crowd workers. Yet, 16.8\% come from app-specific pairs where annotators explore the app before submitting tasks, and not category-clustered pairs. 
This illustrates the need to capture task feasibility, as someone familiar with an app can still pose infeasible requests.
\begin{table}[t]
\renewcommand{\arraystretch}{0.9}
    \centering
    \begin{tabular}{|c|c|c|c|c|c|}
    \hline
      \multirow{2}{*}{\#} & \multirow{2}{*}{Feasible}  &  \multicolumn{3}{c|}{Infeasible} & \multirow{2}{*}{Total} \\
      \cline{3-5}
      & & I & U & P & \\
      \hline
      Demos & 3,323 & 894 & 155 & 295 & 4,667 \\
      \hline
      F/U Qs & 229 & 372 & 154 & 236 & 991 \\
      \hline
    \end{tabular}
    \caption{Task demo breakdown for task feasibility and follow up questions.}
    \label{tab:possible}
    \vspace{-4mm}
\end{table}

Table~\ref{tab:possible} breaks down the number of feasible and infeasible tasks and the reasons for why a task is not possible. These reasons correspond to the multiple choice options available in the demo post survey: (I) the action cannot be completed in the app, (U) the action is unclear or under-specified, and (P) the task seems to be possible, but they cannot figure out how to perform it or other tasks need to be completed first. 
Table~\ref{tab:possible} also includes the number of follow up questions collected for each scenario. 

\smallskip
\noindent\textbf{Task Demonstrations} We collect up to ten demos per task and find the average time spent performing a task demo to be about one minute, varying between categories by at most 44 seconds.
The average number of screens/views visited (\ie, number of actions taken to complete a task) is 14. Separating by feasible versus infeasible tasks, we obtain an average of 10 and 22 views visited, respectively. 
\vspace{-4mm}
\section{Experimental Setup}
As MoTIF's samples contain the natural language task, demonstration, binary feasibility labels, multiclass subclass labels for infeasible tasks, and follow up questions, many research areas can be explored. For now,
we provide baseline results for feasibility prediction.
MoTIF contains nearly 4.7k demos, and we reserve 500 for testing.
We propose a simple Multi-Layer Perceptron baseline with two hidden layers of size 512 and 256 for the binary feasibility classification task. Note that these results provide an upper bound on performance, as input task demos can be considered the ground truth exploration needed to determine feasibility, as opposed to a learned agent's exploration.

We perform ablations of the natural language task (T) with various view hierarchy and app screen representations in Table~\ref{tab:feas}. 
We also explore how to aggregate features over time steps in a task demo; \ie, do we average (Avg), concatenate (Cat), or take the last hidden state of an LSTM.  We cap time steps included to 20, as about 80\% of MoTIF's demos are completed within 20 steps. We report F1 score, with `infeasible' considered the positive class, as we care more about correctly classifying tasks that are infeasible, than misclassifying tasks that are feasible. We found the F1 score to consistently be zero using the first, midpoint, last, or all three time steps, confirming the need to include the exploration as input, as MoTIF's task uncertainty is more nuanced than determining relevancy. We do not include these results in Table~\ref{tab:feas} due to space.

In-vocabulary text and view hierarchy words are represented with FastText embeddings and the rest randomly initialized, with fine-tuning allowed during training. For the view hierarchy, we ablate over the element text (ET), IDs (ID) and class labels (CLS). The average embedding is used for both the input task and view hierarchy text.
We also use Screen2Vec~\cite{screen2vec}, a semantic embedding of the view hierarchy that uses no visual input, which represents each view using a GUI, text, and layout embedder.
For visual representations of the app screen, we obtain ResNet152~\cite{resnet} features for the standard ten crops of each app image and average crop features per screen. We also include icon features obtained from a CNN trained to perform icon classification by~\citet{designsemantics}.
\begin{table}[t]
\renewcommand{\arraystretch}{0.9}
\setlength{\tabcolsep}{3.5pt}
    \centering
    \begin{tabular}{|l|c|c|c|}
    \hline
       Features
       & Cat & Avg & LSTM \\
       \hline
     \textbf{(a) View Hierarchy} & & & \\
     T + ET & 33.8 & 16.3 & 27.6 \\
      T + ET + ID & 32.4 & 14.1 & 26.8 \\
     T + ET + ID + CLS & 27.3 & 15.2 & 34.3 \\
     T + Screen2Vec & 25.2 & 23.8 & \textbf{37.3} \\
      \hline
      \textbf{(b) App Screen} & & & \\
      T + ResNet & 14.9 & 6.3 & 31.2\\
     T + Icons & 17.8 & 0.0 & 19.6 \\
      \hline
      \textbf{(c) Best Combination} & & & \\
      T + Screen2Vec + ResNet &\textbf{35.0} & \textbf{36.9} & 37.0\\
      \hline
    \end{tabular}
    \caption{Task feasibility F1 score using a simple Multi-Layer Perceptron. We provide an ablation over input features and how features are aggregated over time.}
    \label{tab:feas}
    \vspace{-4mm}
\end{table} 
\vspace{-1mm}
\section{Results}
\vspace{-1mm}
Comparing the first row of Table~\ref{tab:feas} (a) which only includes view hierarchy text elements to row two and three in which element ID or class information is included, there is a performance trend that less is more. The (T + ET) input features outperform the (T + ET + ID) and (T + ET + ID + CLS) variants when concatenating or averaging over time. However, the LSTM representation of (T + ET + ID + CLS) results in the best F1 score across rows one to three, suggesting that all element information may be helpful when features are aggregated optimally. Maximal performance is obtained with Screen2Vec view hierarchy features when time steps are aggregated with an LSTM, and its performance when features are averaged over time is higher than all other view hierarchy ablations, demonstrating that Screen2Vec is more robust to aggregation method. 

Next, we ablate over visual features of the app screen.
While icon representations are trained on images from the same domain as MoTIF, they are less effective than ResNet features. The F1 score drops to zero when the average icon feature over time is used, illustrating that an average icon representation does not carry useful information for feasibility classification.  These features were also trained with a smaller, non-residual network, and as a result may be less rich than ResNet features. 

Looking at the various ways of aggregating task demo time steps, concatenating features over time or using the last hidden state of an LSTM generally results in better performance, which suggests that a sequential representation is needed. There is one exception to this: when both Screen2Vec and ResNet features are included ((c) in Table~\ref{tab:feas}), averaging over time outperforms concatenation. This may be a result of nuisance information in the concatenated representation. The LSTM aggregation still outperforms the average representation, which may be due to the forget gate correctly losing unnecessary information over the twenty time steps.

The best results for averaging and concatenating over time are obtained when combining Screen2Vec view hierarchy and ResNet screen features. However, this combination does not outperform the Screen2Vec LSTM representation, which has the highest F1 score across all experiments. This suggests a need for better visual features of non-natural images, as including visual representations should only sustain or improve performance.


\vspace{-1mm}
\section{Conclusion}
\vspace{-1mm}
We introduced MoTIF, a new dataset on Mobile app Tasks with Iterative Feedback that contain natural language commands for actions in mobile apps which may not be feasible. Not only is MoTIF the first to capture this type of task uncertainty for interactive visual environments, but it also contains greater linguistic and visual diversity than prior work, allowing for more research toward robust, reliable, and higher impact vision-language methods. Initial results on the binary feasibility classification task demonstrate there is much room for improvement on the feature representations needed to understand feasibility, as well as better architectures for jointly reasoning about visual and text data.
\section*{Acknowledgements}
This work is funded in part by DARPA and the NSF.
\vspace{-2mm}
\bibliography{anthology,custom}
\bibliographystyle{acl_natbib}

\end{document}